\documentclass[letterpaper, 10 pt, conference]{ieeeconf}  

\IEEEoverridecommandlockouts                              

\usepackage{graphics} 
    \DeclareGraphicsExtensions{.jpg,.pdf,.mps,.png} 
    \graphicspath{{fig/} {./}} 
\usepackage{epsfig} 
\usepackage{amsmath} 
\usepackage{amssymb}  
\usepackage{algorithm,algorithmicx} 
\usepackage{algpseudocode} 
\usepackage{bm} 
\usepackage{flushend} 

\newcommand{\fig}[1]{Fig.~\ref{#1}}

\newcommand{\tab}[1]{Table~\ref{#1}}

\newcommand{\eq}[1]{Eq.~(\ref{#1})}

\def\epsgaiji#1{\leavevmode\kern-0.025zw\raise-.37zh\hbox{%
  \epsfile{file=#1,width=1.05zw}}\kern-0.025zw}
\newcommand{\MARU}[1]{{\ooalign{\hfil#1\/\hfil\crcr\raise.167ex\hbox{\mathhexbox20D}}}}

\usepackage{array}

\usepackage{siunitx} 
\usepackage{booktabs} 
\usepackage{multirow} 
\usepackage{tabularx}    
\usepackage{stfloats}    

\usepackage{pgfplots}
\pgfplotsset{compat=newest}
\usetikzlibrary{plotmarks}
\usetikzlibrary{arrows.meta}
\usepgfplotslibrary{patchplots}
\usepackage{xcolor}
\usepackage[hyphens]{url}
\usepackage[hidelinks]{hyperref} 

\usepackage{grffile}
\pgfplotsset{plot coordinates/math parser=false}
\newlength\fwidth
\newlength\fheight

\usepackage{cite}

\title{\LARGE \bf
LIMBERO: A Limbed Climbing Exploration Robot\\Toward Traveling on Rocky Cliffs
}

\author{Kentaro Uno$^{1\dag}$, Masazumi Imai$^{1\dag}$, Kazuki Takada$^{1}$, Teruhiro Kataonami$^{1}$, Yudai Matsuura$^{1}$,\\Antonin Ringeval-Meusnier$^{2}$, Keita Nagaoka$^{1}$, Mikio Eguchi$^{1}$, Ryo Nishibe$^{1}$, and Kazuya Yoshida$^{1}$ 
\thanks{$^{*}$This work is supported by JSPS KAKENHI Grant Number JP23K13281 and JP25KJ0592.}
\thanks{$^{1}$The authors are with the Space Robotics Lab. (SRL) in Department of Aerospace Engineering, Graduate School of Engineering, Tohoku University,
Sendai 980-8579, Japan.
}%
\thanks{$^{2}$The author is with the National Institute of Applied Sciences of Toulouse, 31400 Toulouse, France.}%
\thanks{\textit{$^\dag$These authors contributed equally to this work. Corresponding author is Kentaro Uno} {(Email: {\tt unoken@tohoku.ac.jp})}.}}

\begin{document}

\maketitle
\thispagestyle{empty}
\pagestyle{empty}


\begin{abstract}
In lunar and planetary exploration, legged robots have attracted significant attention as an alternative to conventional wheeled robots, which struggle to traverse rough and uneven terrain. To enable locomotion over highly irregular and steeply inclined surfaces, limbed climbing robots equipped with grippers on their feet have emerged as a promising solution. In this paper, we present LIMBERO, a 10\;kg-class quadrupedal climbing robot that employs spine-type grippers for stable locomotion and climbing on rugged and steep terrain. We first introduce a novel gripper design featuring coupled finger-closing and spine-hooking motions, tightly actuated by a single motor, which achieves exceptional grasping performance ($>$150\;N) despite its lightweight design (525\;g). Furthermore, we develop an efficient algorithm to visualize a geometry-based graspability index on continuous rough terrain. Finally, we integrate these components into LIMBERO and demonstrate its ability to ascend steep rocky surfaces under a 1\;G gravity condition, a performance not previously achieved yet for limbed climbing robots of this scale. 
\end{abstract}



\section{Introduction} 
Climbing locomotion capability is essential for substantially expanding explorable landscapes in nature. Particularly in space exploration, numerous robotic probes have been launched to the Moon, planets, and asteroids to investigate geological and biological histories, as well as potential resources for near-future sustainable utilization~\cite{spaceRoboticsReview}. 
In such previous space missions, most robots have employed wheels for mobility, considering their high efficiency and traversability on sandy smooth terrain.
However, scientifically valuable regions on the Moon and planets are characterized by highly uneven and steep terrain, exemplified by lunar vertical pits called ``skylights'' and Martian outcrops, which remain largely unexplored due to the limited accessibility of conventional wheeled robots.
\begin{figure}[t]
  \centering
  \includegraphics[width=\linewidth]{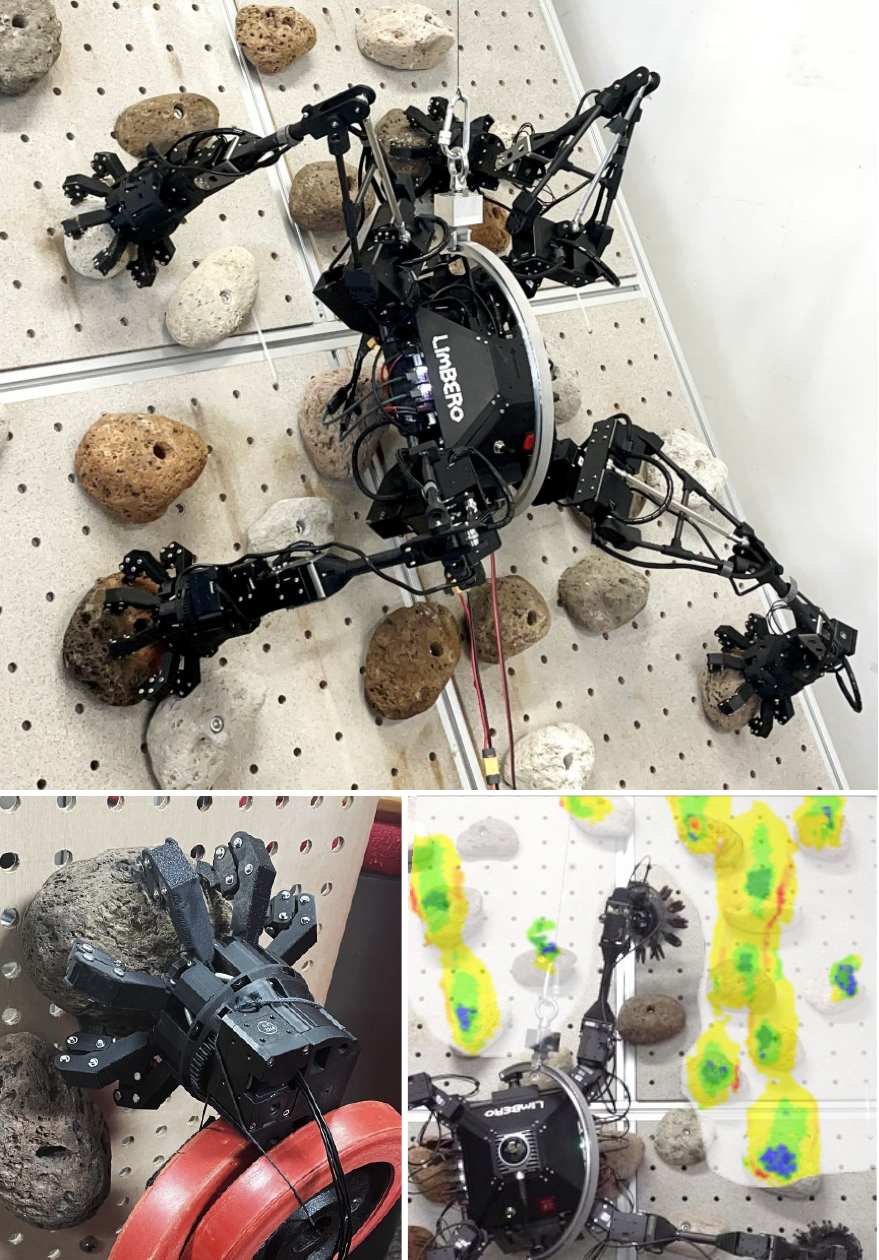}
  \caption{LIMBERO: A Limbed Climbing Exploration Robot. LIMBERO's spine gripper can hold 15\;kg load in Earth gravity (bottom left). Developed algorithm can visualize graspability score in the irregular terrain map (bottom right).}
  \label{fig:LIMBERO_diag}
\end{figure}
To travel across such challenging terrains, many other types of mobile robots
have been investigated~\cite{thoesen2021planetary}, and legged robots, which offer superior traversability on rough terrain due to their ability to jump, climb, and select appropriate contact points, have been getting a spotlight as promising alternatives to conventional wheeled robots for advanced planetary exploration missions~\cite{anymalForPlanetaryExploration,kolvenbach2019towards,spaceclimber,giorgio2023ClimbingANYmal}.

Furthermore, legged robots equipped with spine grippers on their feet, which enable the robot to adhere to rough surfaces, can achieve cliff-climbing capability as well as stable locomotion in low-gravity environments without unintended floating or drifting~\cite{warley2023ramp}.
Many legged climbing robots have been prototyped to date (see \tab{tab:comparison}), demonstrating their gaits and climbing performance on rough surfaces such as rocks, concrete walls, and climbing handholds. For instance, LEMUR~3~\cite{parness2017lemur} (weight: 35\;kg), developed by NASA JPL, demonstrated promising traversability on rough terrain through ceiling-clinging experiments and outdoor cliff-climbing tests. It is equipped with grippers on all four feet, each bundling hundreds of claws known as Microspines. Such a design, with an excessive number of claws, ensures that a sufficient number of spines make contact even on irregular surfaces; however, it results in a redundant system, making the overall robot considerably large in size and mass.
To address this, several lightweight climbing robots have been developed by designing smaller grippers and legs with a reduced number of Microspine units and degrees of freedom (DOFs):
RockClimbo~\cite{RockClimbo2023} successfully climbed vertical surfaces in a simulated environment by grasping surface protrusions with its four-fingered grippers, each containing multiple Microspines;
LORIS~\cite{nadan2024loris} employs a two-fingered passive gripper that hooks onto the surface using the robot’s own weight, enabling climbing on a variety of surfaces, from smooth rock walls to rugged terrain;
HubRobo~\cite{uno2021hubrobo}, a compact and lightweight legged climbing robot equipped with passive spine grippers~\cite{nagaoka2018passive}, demonstrated climbing on sloped rough terrain in a simulated Martian-gravity test field.
SCALER~\cite{tanaka2025scaler}
features a gripper consisting of two fingers with multiple claws and a linkage mechanism, enabling it to grasp surface protrusions for vertical wall climbing and ceiling locomotion under Earth gravity.
ReachBot~\cite{chen2024locomotion}, designed for Martian cave exploration, differs from conventional legged robots in that it uses extendable booms as limbs instead of articulated legs, realizing a large workspace.
\begin{table*}[bthp]
\caption{Comparative summary of the state-of-the-art limbed climbing robots (ordered by weight).}
\label{tab:comparison}
\centering
\begin{tabular}{c|c|cc|rlc|cr}
    \toprule
    \multirow{2}{*}{\bf{Robot}} & \hspace{-2mm}\multirow{2}{*}{\centering\bf{Mass}}\hspace{-2mm} & {\centering\bf{Num. of}} & {\centering\bf{Active DOF}} & \multicolumn{2}{c}{\bf{\hspace{-10mm}Gripper}} & \hspace{-8mm}{\centering\bf{Grasping Force}}\hspace{-2mm} & \multirow{2}{*}{\bf{Terrain}} & \multirow{2}{*}{\centering\bf{Gravity}} \\
    & & {\centering\bf{Limbs}} & \hspace{-3mm}{\centering\bf{per Limb}}\hspace{-2mm} & {\centering\bf{Actuation}} & {\centering\bf{Gripping Principle}$\;^{\ddag1}$} & \hspace{-2mm}{\centering\bf{per Gripper}}\hspace{-2mm} & &\\
    \midrule
    LEMUR 3~\cite{parness2017lemur}  & 35\;kg & 4 & 7  & Active & Radial \emph{hooking} & \bf{150~N} & Rocky terrain & 3/8\;G \\ 
    ReachBot~\cite{chen2024locomotion} & 18\;kg & 8 & 3 & Active & Radial \emph{biting} and \emph{hooking} & 22~N $^{\ddag2}$ & Rocky terrain & 3/8\;G  \\
    \bf{LIMBERO}                    & 9.7\;kg & 4 & 4 & Active & Radial \emph{biting} and \emph{hooking} & \bf{150~N} & Rocky terrain & 1\;G \\
    SCALER~\cite{tanaka2025scaler} & 9.6\;kg & 4 & 6 & Active & Two-fingered \emph{biting} & $>$\;30\;N $^{\ddag2}$ & Artificial holds & 1\;G \\ 
    MARCBot~\cite{MARCBot2024} & 4.8\;kg & 4 & 5 & Active & Radial \emph{biting} and \emph{hooking} & 49.0~N & Rocky terrain & 1\;G \\
    RiSE~\cite{spenko2008biologically} & 3.8\;kg & 6 & 2 & Passive & Directed \emph{hooking} & $>$\;10\;N $^{\ddag2}$ & Flat walls & 1\;G \\ 
    RockClimbo~\cite{RockClimbo2023} & 3.5\;kg& 4  & 4 & Active & Radial \emph{biting} and \emph{hooking} & 23.4~N & Rocky terrain & 2/3\;G \\ 
    LORIS~\cite{nadan2024loris}      & 3.2\;kg & 4 & 3 & Passive & Directed \emph{hooking} & $>$\;10\;N $^{\ddag2}$ & Rocky terrain & 1\;G \\ 
    HubRobo~\cite{uno2021hubrobo}    & 3.0\;kg & 4 & 3 & Passive & Radial \emph{biting} & $>$\;9\;N~\cite{uno2020nonperiodic} & Artificial holds & 3/8\;G \\ 
    CLIBO~\cite{sintov2011design}    & 2.0\;kg & 4 & 4 & Passive & Directed \emph{hooking} & $<$\;19.6\;N & Flat walls & 1\;G \\ 
    \bottomrule 
\end{tabular}
\begin{minipage}[]{.08\linewidth}
\end{minipage}
\begin{minipage}[]{.70\linewidth}
\vspace{-3mm}
\footnotesize{$^{\ddag1}$\;In this study, \emph{biting} refers to the event in which a spine is pressed normally onto the surface by the gripper’s finger-closing motion, whereas \emph{hooking} is achieved through tangential movement of the spine, as illustrated in the right-hand images.}\\
\footnotesize{$^{\ddag2}$\;Estimated value based on the reported experiments. The calculation is simply as follows: (The total weight of the robot and payload) $\times$ 9.81 / (The number of grippers supporting the robot).}
\end{minipage}
\begin{minipage}[]{.265\linewidth}
\raggedleft
\includegraphics[width=\linewidth]{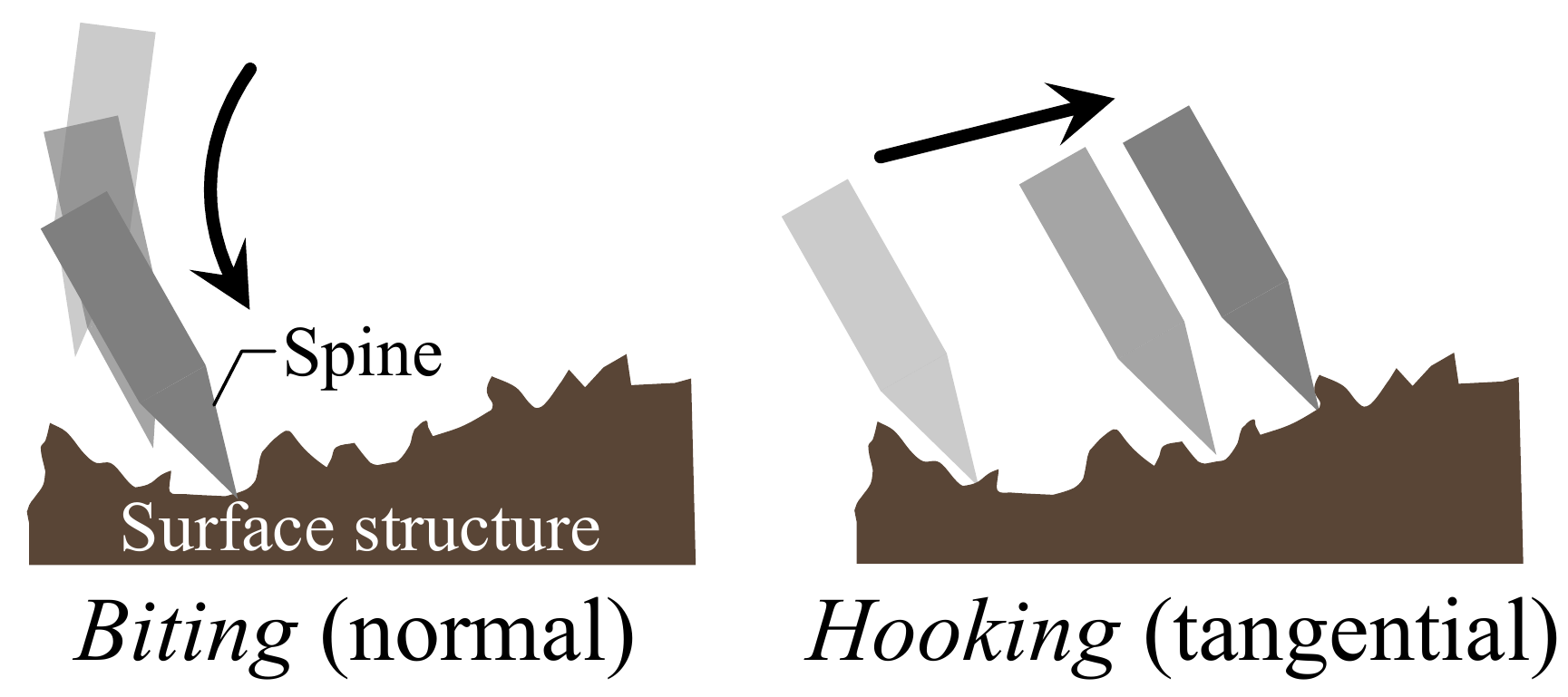}
\end{minipage}
\vspace{-3mm}
\end{table*}
While many legged rock-climbing robots have been developed, the majority of them feature minimal and compact designs, which limits their practical applications. Among them, LEMUR~3 is the only example showing robust climbing capability while supporting a relatively large mass; however, its demonstration was performed under reduced-gravity conditions. Consequently, deploying climbing robots in more rigorous scenarios requires significant technical improvements, particularly in the spine-type gripping mechanism.

To tackle this challenge,
we designed a novel spine-type gripper that maximizes the contact force of each spike by enabling both \emph{biting} and \emph{hooking} motions, resulting in exceptional gripping performance despite its compact size. We then introduce a 10\;kg-class limbed climbing robot, named LIMBERO (LIMBed cLIMBing Exploration RObot), equipped with the grippers. LIMBERO demonstrated a 1.4\;kg payload capacity and stable climbing capability while supporting its own weight under full Earth gravity conditions. To the authors' best knowledge, this work represents the first demonstration of a relatively large-scale climbing robot on rocky terrain. Furthermore, to maximize the gripper's performance, we developed an efficient algorithm to extract suitable graspable geometries from the terrain, which has been open-sourced. Finally, we discuss lessons learned for achieving more stable deployment under more rigorous conditions and for improving operational efficiency.
The key contributions are highlighted as follows:
\begin{itemize}
    \item We present a novel gripper design, featuring coupled finger-closing and spine-hooking motions, tightly actuated via a single motor with a reduction gear mechanism, achieving outstanding grasping performance ($>$150\;N).
    \item We introduce an efficient algorithm to visualize the geometry-based graspability of each local point on continuous rough terrain.
    \item We integrate these components into LIMBERO and demonstrate its ability to ascend steep rocky terrain.
\end{itemize}

\section{Spine Gripper Design}
\begin{figure}[t]
  \centering
  \includegraphics[width=.9\linewidth]{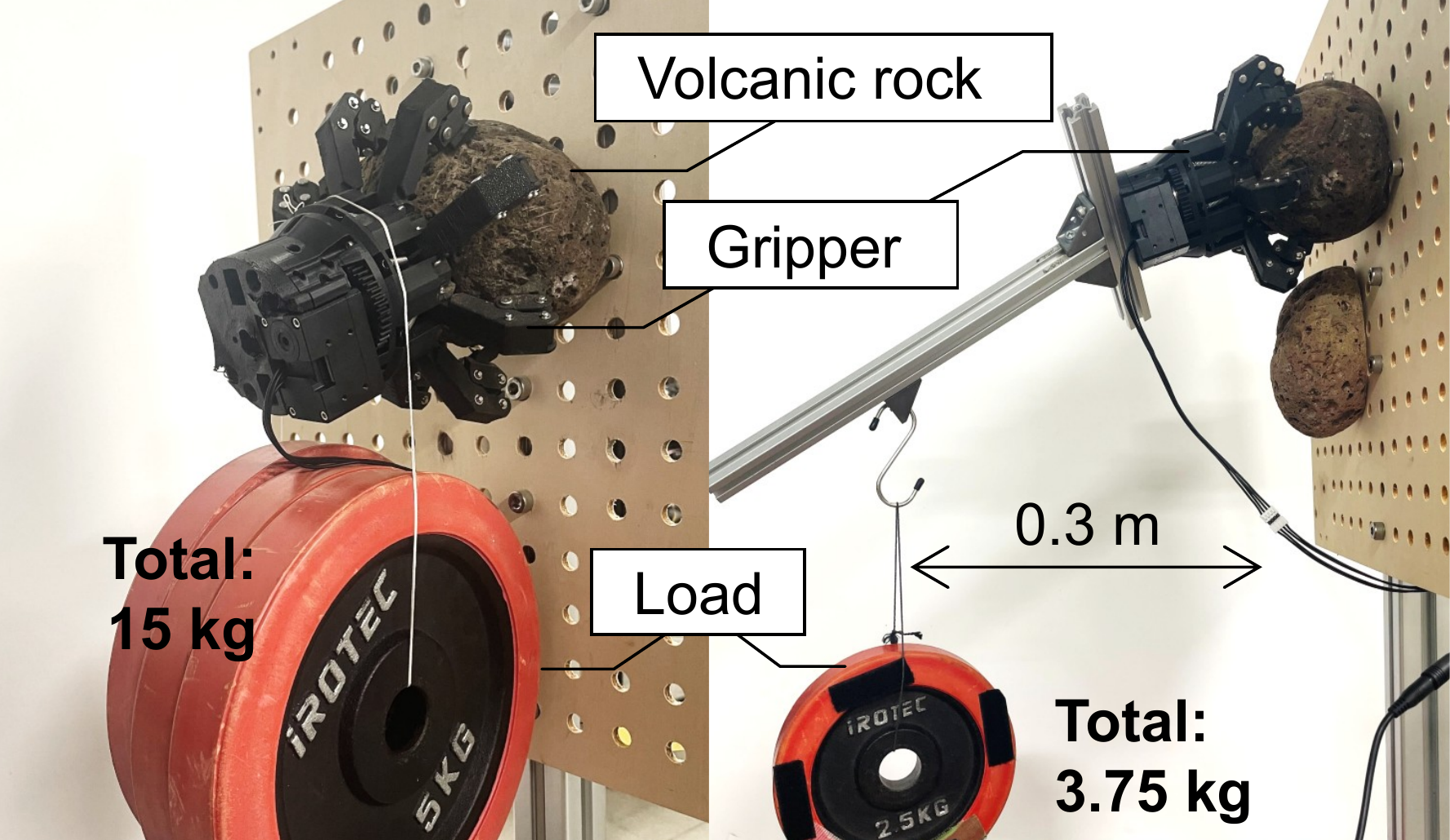}
  \caption{Grasping performance by the developed gripper.}
  \label{fig:gripper_performance}
\end{figure}
In short, the practicality of a climbing robot dominantly depends on the performance of its gripper. LIMBERO's gripper was designed to achieve exceptionally strong engagement with rocky surfaces, enabling it to support the robot’s body along with a payload of several kilograms under Earth gravity, thereby facilitating applications beyond space exploration. The latest version of our gripper successfully supports a load of 15\;kg and withstands an external moment of 10\;Nm, despite its lightweight design (gripper mass: 0.525\;kg), while grasping a volcanic rock (pumice) on a vertical wall (see \fig{fig:gripper_performance}). This result indicates that even a single gripper can adequately support the robot’s weight. Such excess performance is intentional and crucial for providing a sufficient safety margin in case one or more grippers experience unexpected detachment.

\subsection{Principle}
Through numerous prototyping and iterative testing, we found that substantially enhancing the performance of spine-type grippers requires coupling the \emph{biting} and \emph{hooking} actions of the spikes. Here, \emph{biting} refers to a spine penetrating the rocky surface along the normal direction, while \emph{hooking} is achieved via a tangential scratching motion (see \tab{tab:comparison}). Several previous approaches employed only one of these principles, which suffices to support relatively light weights ($<$4\;kg)~\cite{spenko2008biologically,nadan2024loris,uno2021hubrobo,nagaoka2018passive,sintov2011design}, and implementing such coupled motions with a passive gripper is generally challenging.
The LEMUR series' Microspine gripper~\cite{parness2017lemur,parness2013gravity} relies solely on the hooking principle; nevertheless, by incorporating a large number of spines, it successfully achieves the highest grasping force among state-of-the-art systems. However, the overall design is bulky and highly redundant.
SCALER's two-fingered end-effector~\cite{tanaka2025scaler}, called the GOAT gripper,
is another example showcasing promising gripping performance by exploiting only the biting principle. It strongly pinches surface protrusions, enabling the robot to climb vertical walls and traverse ceilings under Earth's gravity; however, due to its pinching nature, it struggles to grasp relatively smooth or gently rough terrain.
On the other hand, RockClimbo~\cite{RockClimbo2023}, followed by MARCBot~\cite{MARCBot2024}, employed semi-passively actuated grippers, called SPASAS grippers, which enable both biting and hooking motions via a single cable-driven actuator. These grippers achieved two to five times greater engagement force than passive grippers that utilize only a single principle.
\begin{figure}[t]
  \centering
  \includegraphics[width=\linewidth]{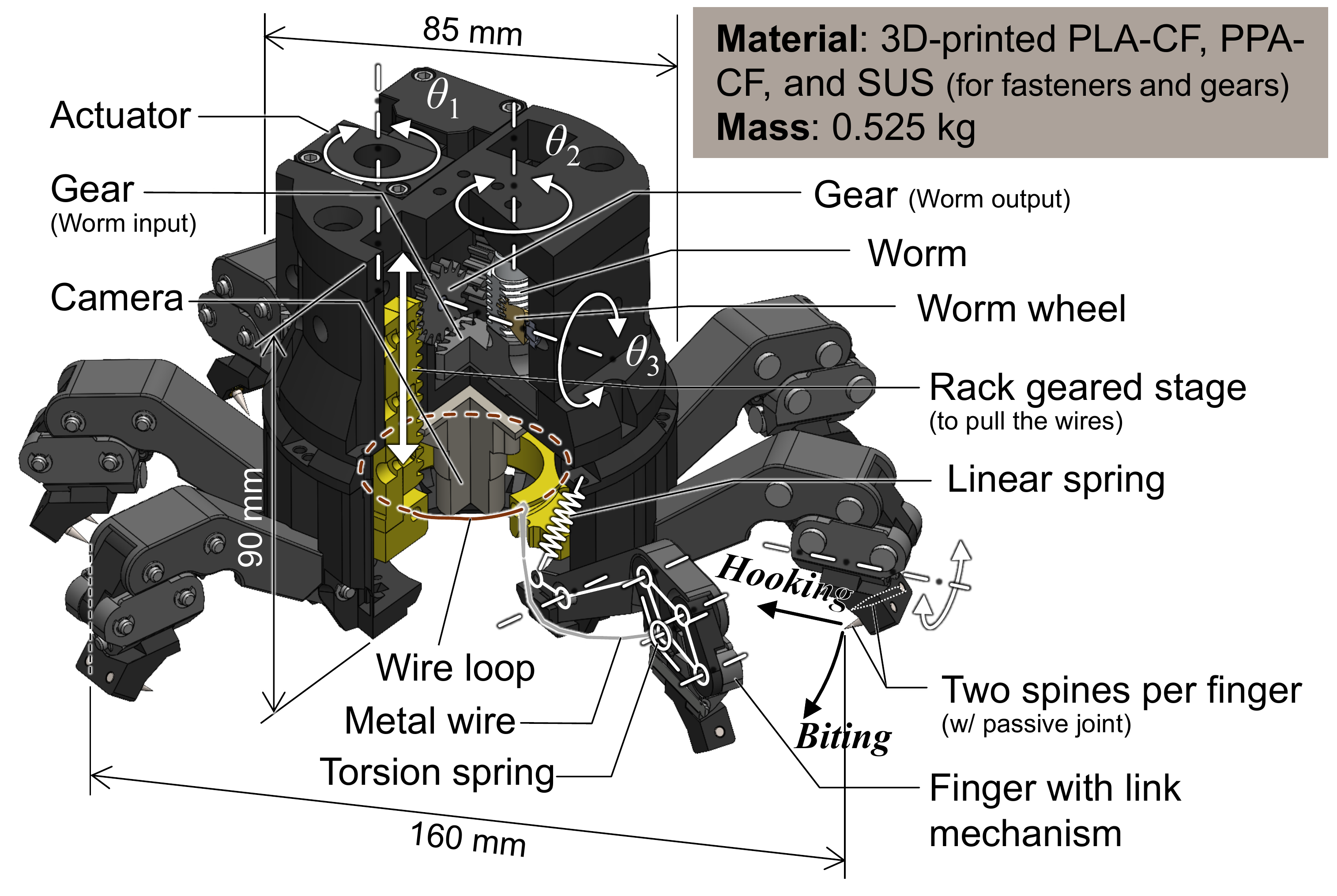}
  \caption{Gripper design: a servo motor drives a worm and two worm wheels, which linearly move the stage (yellow) to tighten or loosen the metal wires, thereby opening and closing the gripper.}
  \label{fig:gripper_design}
\end{figure}
\subsection{Mechanism}
Our developed gripper employs a linkage mechanism that enables both \emph{biting} and \emph{hooking} motions using a single actuator. The detailed mechanism is illustrated in \fig{fig:gripper_design}. Each of the eight radially arranged fingers is equipped with two spines, yielding a total of 16 spines per gripper. The dual spines are mounted on passive roll joints, providing additional adaptability similar to that of previously proposed novel Microspine grippers~\cite{parness2017lemur,parness2013gravity}. This underactuated flexibility is crucial for increasing the number of spines in contact with the surface, thus distributing the contact forces more evenly across the spines. The finger-closing and spine-scratching motions are actuated by pulling and releasing metal cables connected through another wire loop to a vertically translating stage driven by a geared mechanism: input motor ($\theta_1$) -- worm ($\theta_2$) -- worm wheel ($\theta_3$). The resulting reduction ratio is approximately 25:1, enabling the actuator to strongly tension the cables, thereby producing a high contact force at each spine. To avoid maintaining the actuator in a stalled state, a worm gear is employed, allowing the actuator current to be cut off while the high-friction mechanical lock holds the position. Each finger incorporates two springs: an extended linear spring at the finger root, which closes the finger, and a compressed torsion spring at the finger tip, which extends it. Consequently, when the cable is pulled, the finger-closing motion---\emph{biting}---always precedes the spine-scratching motion---\emph{hooking}, and the sequence is reversed when the cable is released. It is important to note that the flexibility of the wire loop contributes to balancing the loads among the fingers to some extent. Although this design does not achieve the same level of load-sharing performance as the mechanism employed in~\cite{chen2022reachbot}, it offers a more compact and simpler configuration. It is also noteworthy that the individually assigned springs on each finger are critical for ensuring adaptability to arbitrary terrain shapes, even though the displacement of the pulled cables is identical across all fingers. 

\section{Limbed Climbing Robot Design} 
LIMBERO, a quadrupedal climbing robot, comprises four limbs, each equipped with the developed spine gripper, connected to a central base. The total robot mass is 9.7\;kg, consisting of a 2.8\;kg central body and limbs weighing 1.72\;kg each, including the grippers. The prototype was fabricated primarily using Carbon Fiber Reinforced Plastic (CFRP) tubes and 3D-printed components made of carbon fiber-reinforced Polylactic Acid (PLA-CF) and Polyphthalamide (PPA-CF), while stainless steel (SUS) and aluminum were used only for components requiring high durability (e.g., gears and fasteners). This approach resulted in a lightweight yet robust design.
\subsection{Limb Design} 
\begin{figure}[t]
  \centering
  \includegraphics[width=.97\linewidth]{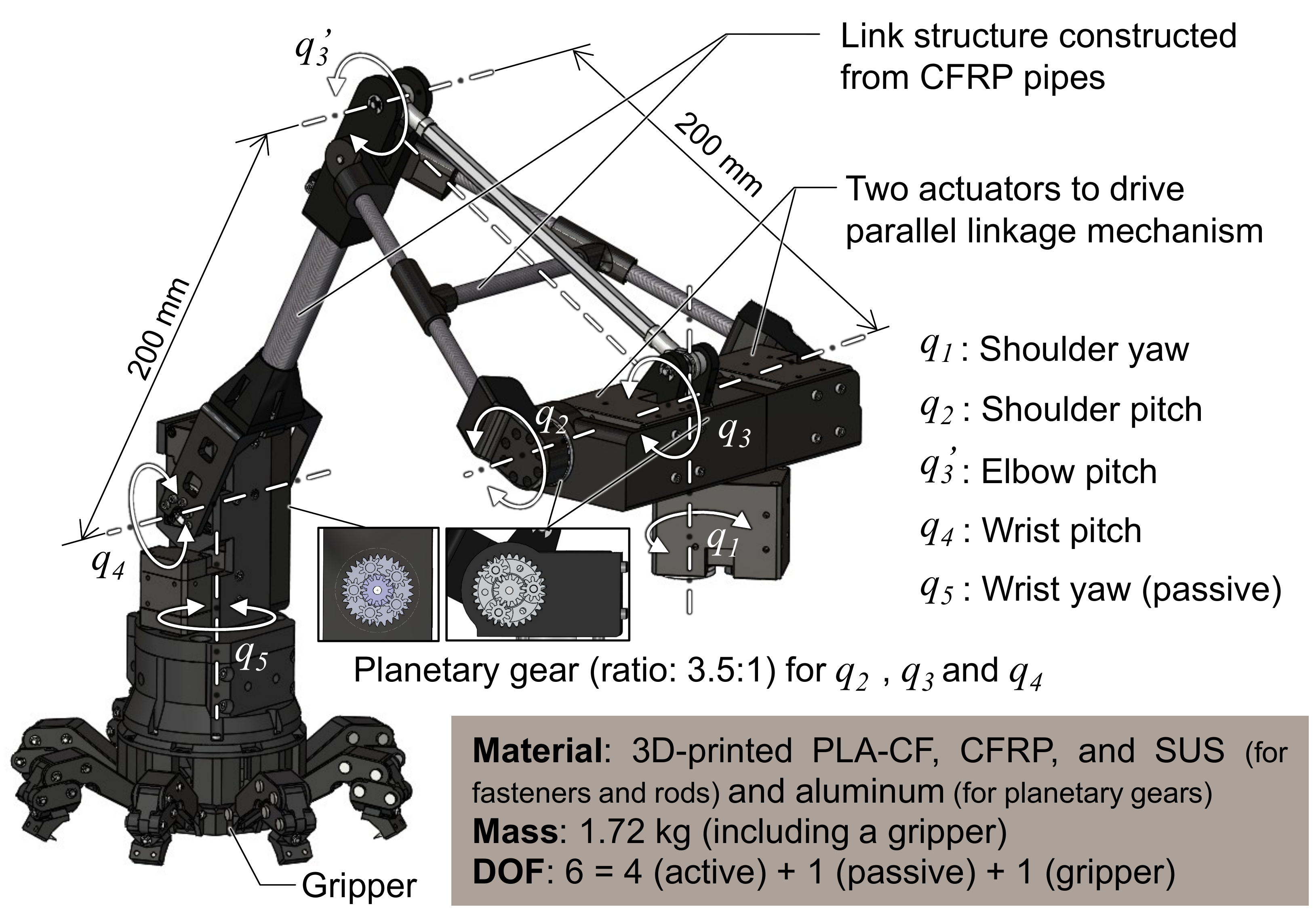}
  \caption{Limb design: three actuators are concentrated at the shoulder via a parallel-linked mechanism, reducing limb inertia while preserving a large workspace. Three pitch joints requiring high torque capacity are equipped with a 3.5:1 planetary gear reduction.}
  \label{fig:limb}
\end{figure}
Since climbing robots rely on their limbs to support themselves and generate motion on the terrain, both limb mass and range of motion are critical design factors. To reduce limb inertia while providing the necessary degrees of freedom for climbing, LIMBERO employs a parallel-linked mechanism (see \fig{fig:limb}). Each limb features four active joints: three translational DOFs of the end-effector, realized by three revolute joints arranged in a yaw--pitch--pitch configuration---commonly referred to as an ``insect-type'' configuration---which has been shown to be superior for climbing in terms of tumble stability and kinematic reachability~\cite{uno2021simulation}; and an additional pitch joint that maintains the gripper palm parallel to the terrain surface, minimizing the number of spines losing contact. To prevent torsional moments at the leg tips when moving the base while all four grippers are engaged, a passive yaw DOF is added at the upper part of the gripper. Consequently, each limb has four active DOFs and one passive DOF. Dynamixel series actuators were selected for all joints, considering their compact size and minimal daisy-chained cabling: three XM540-W270 units drive the yaw--pitch--pitch joints, and one XM430-W350 drives the final pitch joint. To enhance torque capacity, the three pitch joints are equipped with planetary gear reducers with a reduction ratio of 3.5:1 (input: sun gear; output: planetary carrier). All limbs share the same structural design, enabling modularity in both hardware and software.




\subsection{Base Design} 
\begin{figure}[t]
  \centering
  \includegraphics[width=\linewidth]{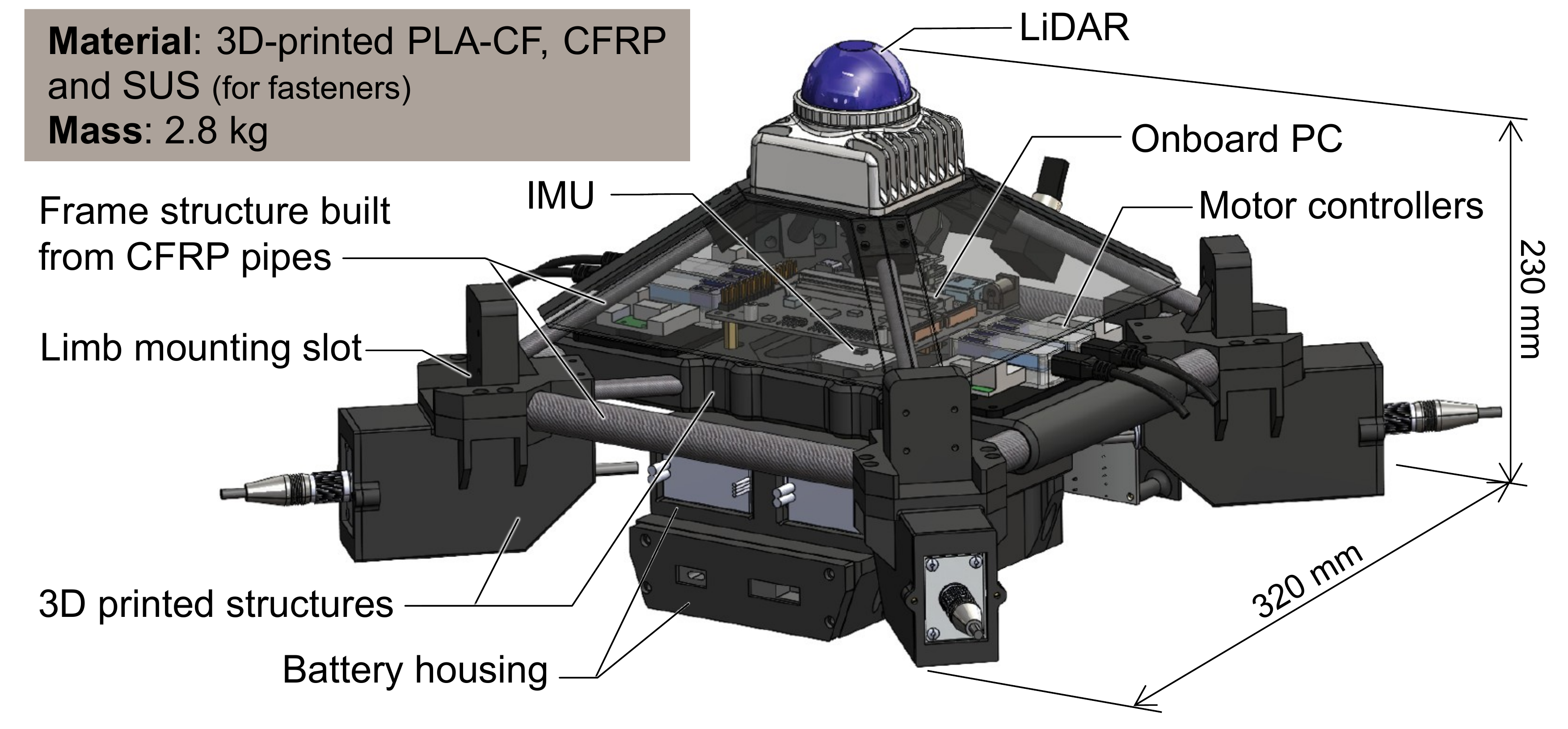}
  \caption{Base design: an axisymmetric structure accomodates electronics, sensors, and batteries.}
  \label{fig:base}
\end{figure}
The base of LIMBERO is designed with a square top view to ensure an axisymmetric structure. It is fabricated using 3D-printed plastic, with CFRP pipes serving as the primary structural components, achieving a lightweight yet highly rigid design. The base houses the main onboard PC (Jetson Orin NX), an IMU (RT-USB-9axisIMU2), and motor controller ICs for the Dynamixel actuators. Its quadrangular-pyramid shape accommodates a 3D LiDAR (Livox Mid-360) on the upper part, providing a wide field of view while minimizing the number of supporting pipes to enhance structural strength. For untethered operation, high-capacity batteries are positioned at the bottom of the base to lower the center of gravity and maintain climbing stability: a 24,000\;mAh Li-ion pack for the control boards, and two 3-cell 5,200\;mAh LiPo packs for the actuators.

\subsection{Software Design}
\begin{figure}[t]
  \centering
  \includegraphics[width=.9\linewidth]{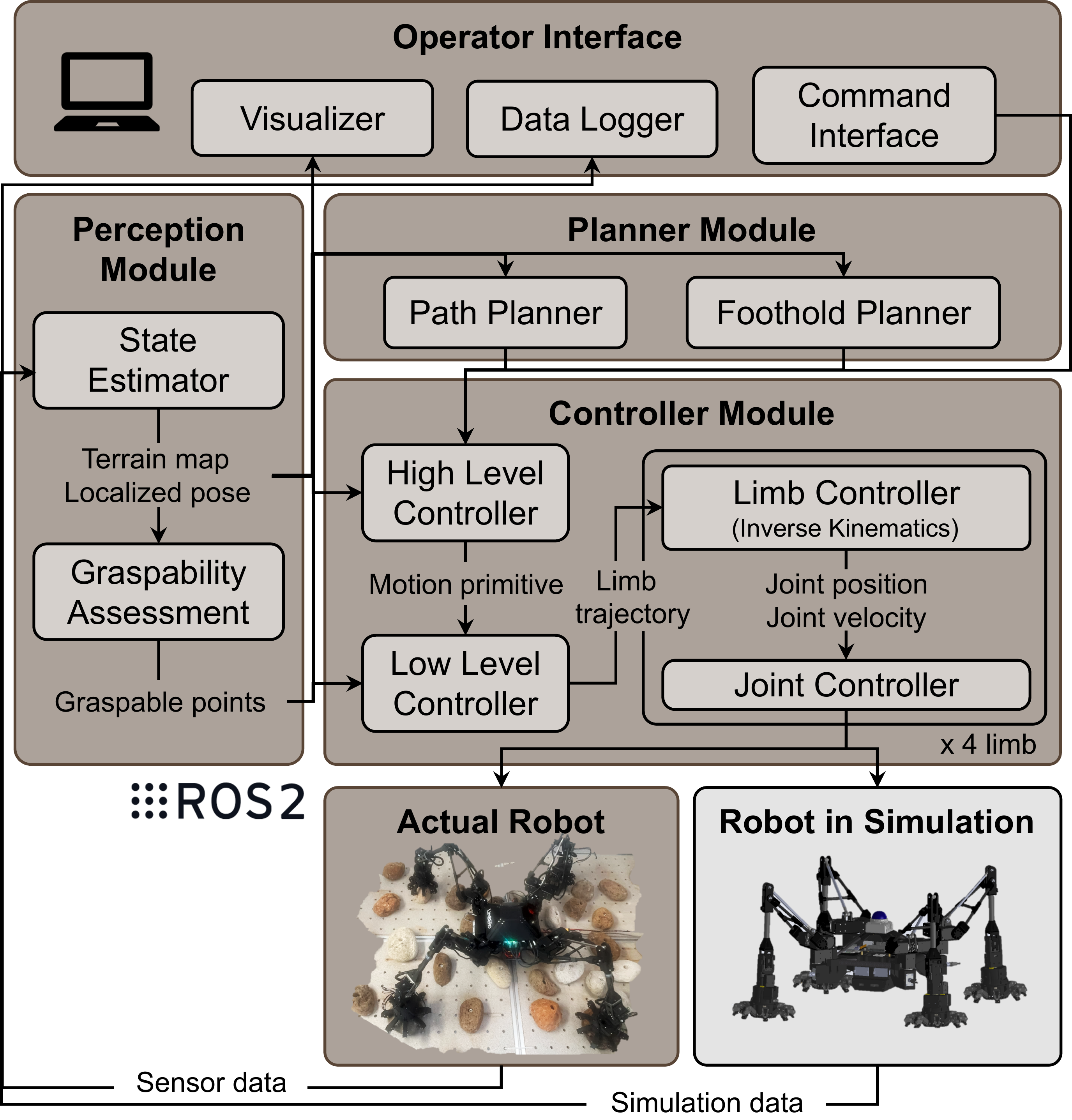}
  \caption{Software architecture implemented for LIMBERO.}
  \label{fig:software}
\end{figure}

LIMBERO’s software is implemented in C++ and Python using ROS\;2~\cite{ros2}, adopting a modular architecture that mirrors the hardware configuration to ensure extensibility. As shown in \fig{fig:software}, the system is organized into three primary functional modules: perception, planner, and controller.
Commands for step planning (e.g., swing limb selection and the next gripping position relative to the robot frame) from the operator or the planner are transmitted to the High-Level Controller (HLC), which generates motion primitives processed by the Low-Level Controller (LLC). Within the LLC, high-level commands---such as movement direction and displacement in the robot coordinate system---are converted into individual limb motions and sent to the Limb Controller. The LLC manages coordinate transformations for each limb to perform control in task space. The Limb Controller computes target joint angles and velocities in joint space using inverse kinematics, which are then sent to the Joint Controller to command the actuators along the desired trajectories. Both controllers operate in parallel for each limb, preventing delays during simultaneous limb movements. 
Sensor data are processed by the State Estimator to compute the robot's pose and the contact states of its end effectors, which are subsequently transmitted to the planner to determine the next actions.

\section{Graspability Assessment}
To maximize the performance of a climbing robot’s gripper, it is essential to select appropriate gripping locations on the terrain. Grasping points can be evaluated based on multiple factors, including surface shape, roughness (friction), stiffness, and fragility. However, fully measuring all these parameters is challenging. While shape and roughness can be sensed visually or predicted from 3D terrain data, stiffness and fragility require contact-based assessment. A previous study demonstrated a method for detecting and selecting grasping points by focusing on the local contact surface between the spine and the rock, choosing candidates from terrain points approximated as spheres based on scanned rock surfaces~\cite{chen2024locomotion}.

In this study, assuming the terrain consists of rough rocky surfaces with micro‑asperities suitable for spine hooking (unlike the overly smooth surfaces of rounded river rocks), we focus on detecting geometrically graspable shapes in a three-dimensional terrain map. The developed source code is publicly available\footnote{The source code is available here: 
\url{https://github.com/Space-Robotics-Laboratory/graspable_points_detection_ros2}
}.

\subsection{Problem Formulation}
Grasping targets should have appropriate size and sharpness to fit within the gripper’s graspable space. The problem can be formulated as identifying proper shapes, i.e., the range of sizes and degrees of sharpness that fit within the gripper’s graspable space, from the terrain point cloud. The algorithm presented here, in essence, scans the 3D terrain geometry using the predefined range of graspable shapes of the gripper.
\begin{algorithm}[t]
    \caption{Geometric Graspability Assessment}
    \label{alg:grasping_target_detection}
    \begin{algorithmic}[1]

        \State {\bf Input:}
        \State $P$: Raw point cloud;
        \State $\Gamma$: Geometric parameters of the robot’s gripper;
        \State $c$: Voxel grid size (mm);
        \State {\bf Output:}
        \State $g$: Graspable score for all the points $p$ in $T$;
        \State $G$: A set of potentially graspable points;
        
        \State {\bf Main function:}
        \State $T =$ {\sc CreateTerrainArray}$(P, c)$;
        \State $M =$ {\sc CreateGripperMask}$(\Gamma, c)$;
        
        \Function{GraspabilityAssessment}{$P$, $\Gamma$, $c$}
        
        \For{all the points $p$ in $T$}
            \State $T^* = $ {\sc ExtractSubsetTerrainArray}$(T,M)$;
            \State $g = $ {\sc ScoreVoxelByComparison}$(T^*,M)$; 
            \If{$g \geq \hat{g}$}
                \State $G \leftarrow p$;
            \EndIf
        \EndFor
        \State \Return $g$, $G$;
        \EndFunction
    \end{algorithmic}
\end{algorithm}
The problem is mathematically formulated as shown in Algorithm~\ref{alg:grasping_target_detection}. The input to the algorithm is raw, unfiltered point cloud dataset: $P$ composed of $n$ spatial points: ${P}_i \in \mathbb{R}^3$, $i \in \{1, ..., n\}$, perceived by the depth sensors of the robot. The algorithm also takes as input a set of geometric parameters of the gripper: $\Gamma$, that will be used to obtain the spatial array called \emph{gripper mask}: $M$. For instance, $\Gamma$ includes the diameter of the palm, the finger length, and the range of the finger joints, i.e., the geometric parameters that define the gripping range of motion (see the source code documentation$^\ddag$ for a more detailed definition of $\Gamma$). The algorithm first outputs the 3D voxel array obtained through preprocessing the input raw point clouds: $T \in \{0,1\}^{i \times j \times k}$, called \emph{terrain array}. Here, ${i,j,k}$ are ${x,y,z}$-directional size of the array, which is determined by the preset value of voxel grid size $c$. 
In the subsequent steps, each voxel in $T$ is associated with its spatial coordinates and, additionally, a graspable score $g \in [0,1]$, which describes the suitability of the respective point.
Using $g$, the algorithm outputs $G \subseteq T$, a subset of the preprocessed point cloud, containing the points with graspability scores exceeding a predetermined threshold $\hat{g}$.
\begin{figure}[t]
  \centering
  \includegraphics[width=\linewidth]{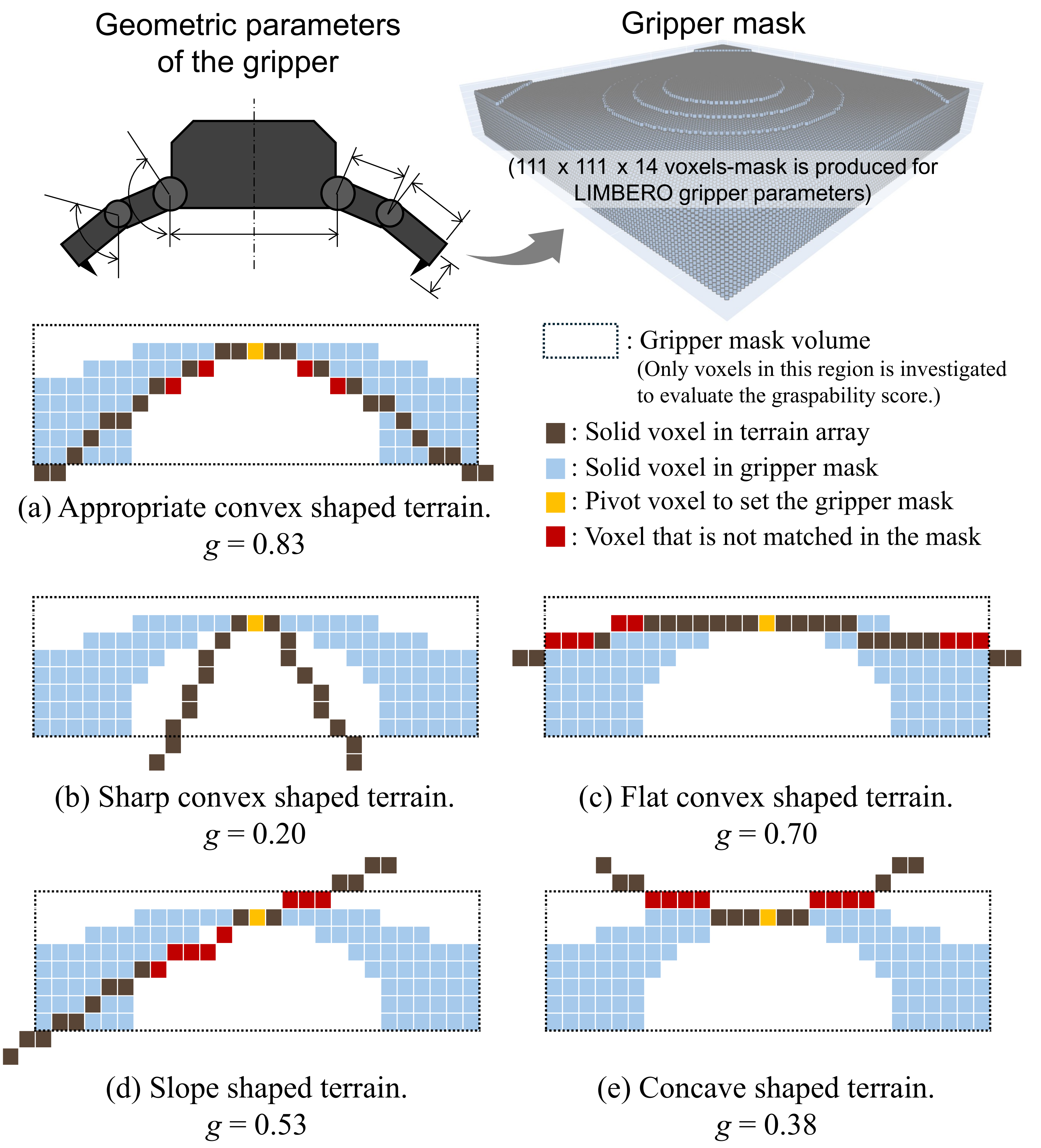}
  \caption{Case study of the gripper mask matching method over various terrain. The terrain array and the gripper mask are illustrated in simplified cross-sectional view.}
  \label{fig:alg_case_study}
\end{figure}
\subsection{Terrain Array}
The \emph{terrain array} is an efficient representation of the three-dimensional shape of the ground surface. The terrain array is defined through the following steps. 
\begin{enumerate}
  \item Coordinate transformation: the coordinates are transformed to the fixed reference frame of the regression plane obtained from the raw point cloud, such that the $z$-axis is aligned with the normal direction of the regression plane.
  \item Occlusion compensation: the occluded area is linearly interpolated by Delaunay triangulation. As a result of this operation, the density of the point cloud is homogenized.
  \item Voxelization and labeling: the homogeneously distributed point cloud space is voxelized, where each voxel is labeled as either solid: 1 or void: 0, depending on whether the voxel contains point cloud data. 
\end{enumerate}
Through this operation, the terrain information, which is originally represented as a point cloud, is reformulated into a simpler description as a 3D array consisting of boolean elements: either 1 or 0, $T \in \{0,1\}^{i \times j \times k}$.


\subsection{Gripper Mask}
The \emph{gripper mask} is the second tool used to perform graspable convexity detection. It is also a 3D array with the same pitch as the terrain array, but it represents the graspable volume of the gripper, which is determined by the gripper design parameters $\Gamma$. While gripper masks of any shape can be designed, for mechanical fingered grippers, the mask is designed based on the angular range of motion of the fingers. For such an axisymmetric fingered gripper for rock grasping, such as HubRobo's \cite{nagaoka2018passive,uno2021hubrobo} or LEMUR robots' \cite{parness2017lemur}, the mask can be designed as a circular truncated cone shape. The voxel located at the top center of the \emph{gripper mask} is referred to as the pivot point (marked in orange in \fig{fig:alg_case_study}) and is used to define the representative position of the mask. 

\subsection{Geometric Graspability Assessment}
\begin{figure}[t]
  \centering
  \includegraphics[width=\linewidth]{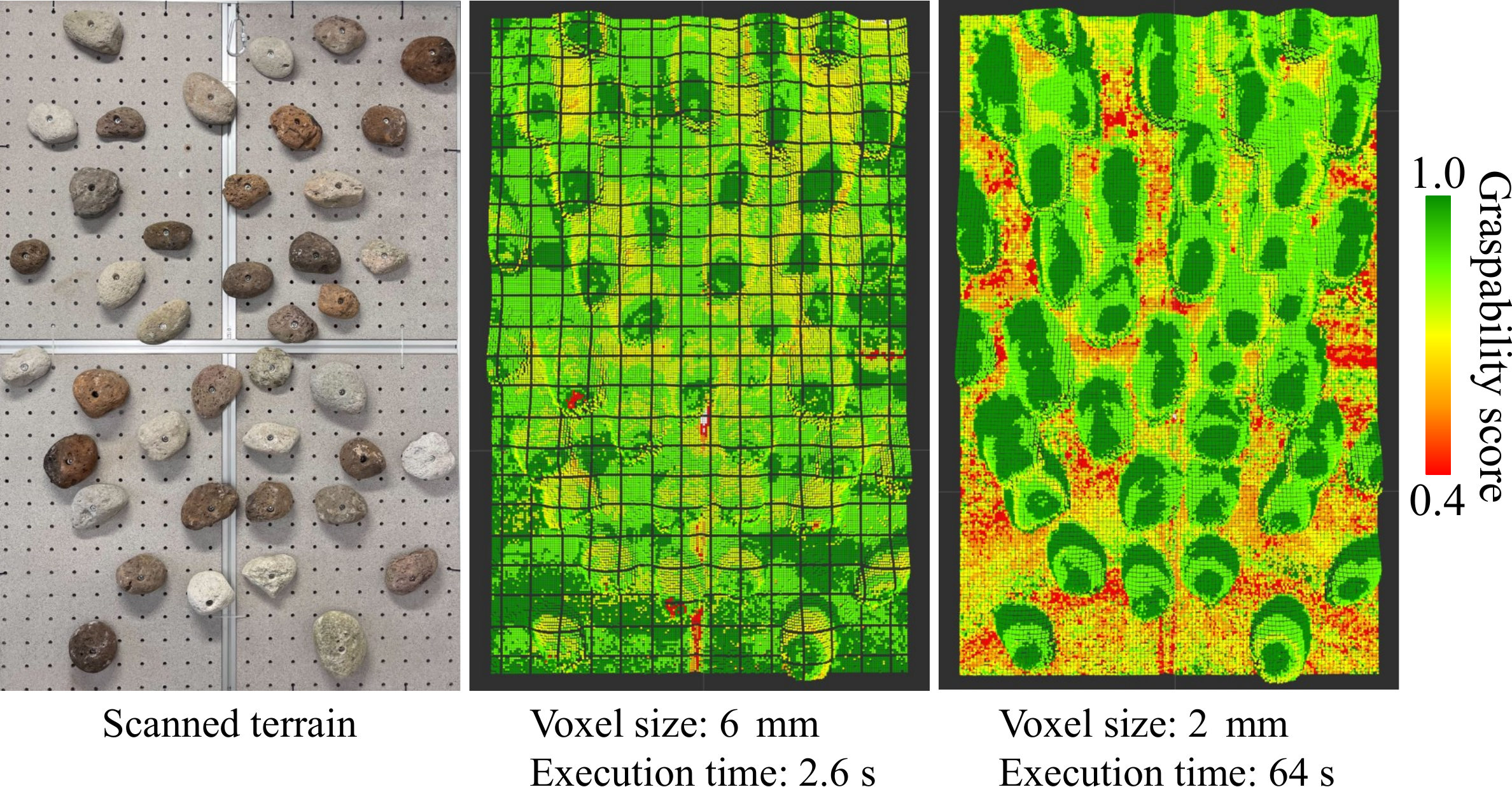}
  \caption{The output of the graspability is an interpolated point cloud, with colors representing the graspability score of each point. Higher-resolution settings for voxelization require longer computation time, but achieve more accurate scoring (right).}
  \label{fig:graspability_scored_map}
\end{figure}
To evaluate the graspability of a point $p$ in the terrain array~$T$, the pivot point of the gripper mask is first positioned at $p$, and then a subset of the terrain array $T^*$, with the same size as the gripper mask $M$ is extracted around $p$. 
In short, the algorithm assesses how similar $T^*$ is to $M$. In the algorithm, for all points $p$ in $T$, 
the loop computes a graspability score $g$, derived from the ratio of the matching quantity between solid voxels in the \emph{terrain array} subset $T^*$ and \emph{gripper mask} $M$, which is simply computed as the inner product of those two same-sized $\{0,1\}$ arrays, and the absolute number of solid voxels in $T^*$, which is also simply computed as the squared Frobenius norm $\left( \| \cdot \|_F \right)$ for a multi-dimensional array (\eq{eq:met1}). To further reduce the computational cost, the algorithm allows a $z$-threshold, such that the computation evaluates only points above the threshold value.
\begin{equation}
	g =  \frac{ M \cdot T^* }{ {\left\| T^* \right\|^{2}_F} } 
	\label{eq:met1}
\end{equation}
\fig{fig:alg_case_study} illustrates the simple case study of the graspable score $g$ for different terrain shapes. For clarity, two-dimensional voxel arrays were examined.

\subsection{Result}
The graspability assessment algorithm was applied to the prepared terrain, where volcanic rocks (pumice) were randomly distributed (see \fig{fig:graspability_scored_map}). As a result, higher scores are observed around the summits of convex geometries, whereas other regions that are locally sloped or concave receive lower scores. This indicates that the algorithm effectively evaluates geometrical graspability across the terrain map. The input data comprised 108,745 raw point cloud points. With a voxel size of 2\;mm and no $z$-threshold, enabling the terrain to be scanned entirely, even in concave regions, the algorithm executed in 64\;s on LIMBERO's onboard PC (Jetson Orin NX). This computational time can be further shortened by using a lower resolution for the voxelization process, which, however, affects the accuracy of graspability scoring.

\section{Experiments}
Several experiments were conducted with LIMBERO. The valuable insights gained from these tests will be essential in guiding future hardware improvements and the development of more advanced, autonomous operations.

All experiments were performed in an indoor climbing test field~\cite{uno2024testfield}.
For the test, on a steeply inclined (60$^\circ$) field, pumice stones were placed randomly, and the robot was powered externally without batteries. First, the terrain was pre-scanned using a LiDAR sensor, and its graspability was evaluated offline (\fig{fig:graspability_scored_map}). Subsequently, the robot was teleoperated by a human operator to climb the rocky surface by selecting regions with high graspability scores.

\subsection{Rocky Cliff Holding Experiment}
\begin{figure}[t]
  \centering
  \includegraphics[width=.8\linewidth]{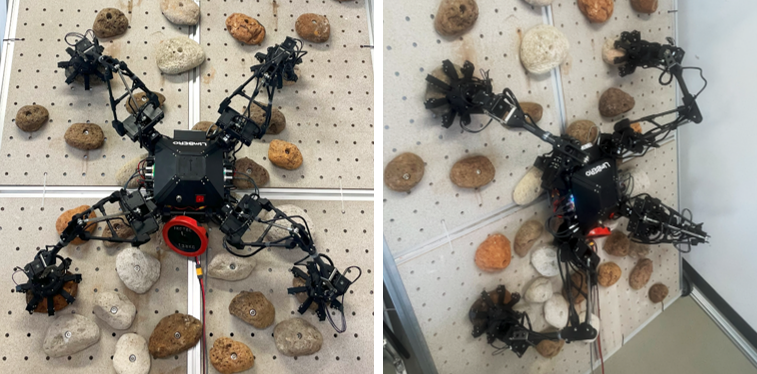}
  \caption{Cliff holding test by LIMBERO. The robot demonstrated 1.4\;kg payload capacity under Earth gravity.}
  \label{fig:cliff_holding_test}
\end{figure}
As a preliminary test to evaluate the basic performance of LIMBERO, we conducted a static cliff-hanging experiment. In the test, the robot stably supported its own body as well as an additional 2.5\;kg load. Accounting for the mass of the omitted batteries, this result indicates a payload capacity of approximately 1.4\;kg under full Earth gravity.

\subsection{Rock Climbing Experiment}
\begin{figure*}[t]
  

  \begin{minipage}[]{.70\linewidth}
  \raggedright
  \includegraphics[height=29.4mm]{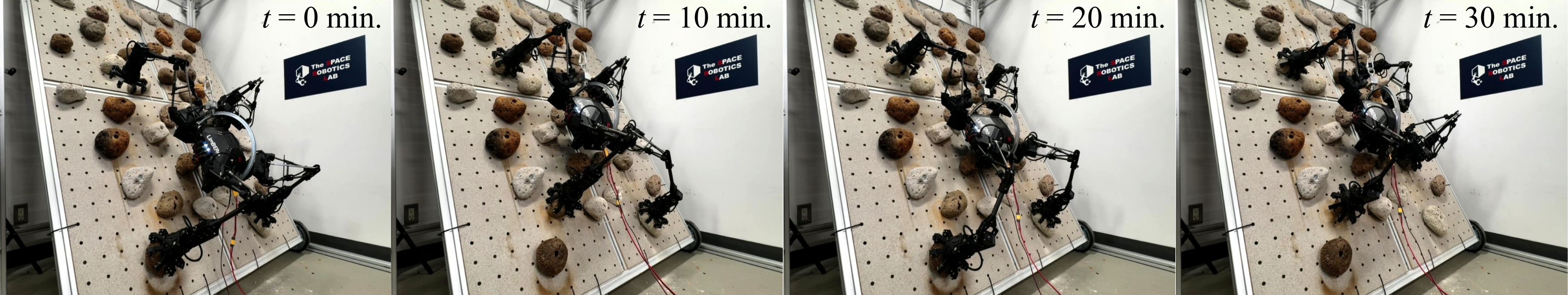}
  \end{minipage}
  \begin{minipage}[]{.30\linewidth}
  \raggedleft
  \includegraphics[height=29.4mm]{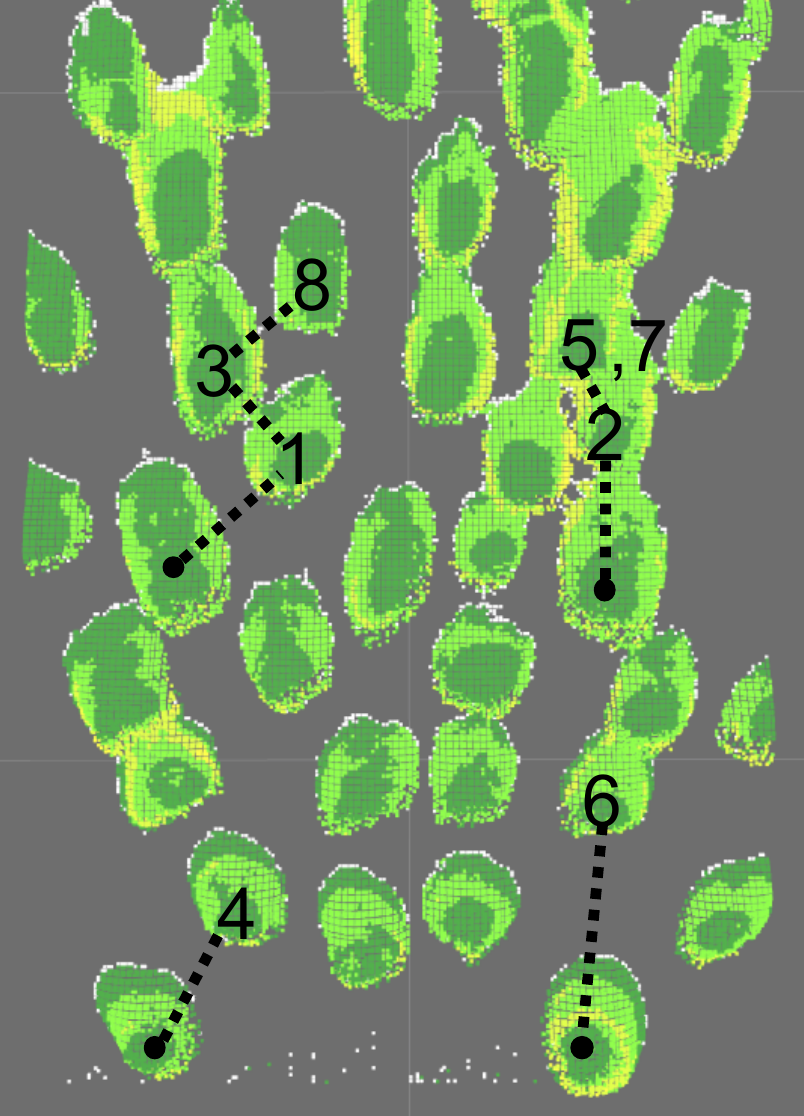}
  \end{minipage} \\
  \begin{minipage}[]{\linewidth}
  \centering 
  \end{minipage} \vspace{-3mm}
  \caption{Rock climbing with LIMBERO under full Earth gravity. The right foothold history indicates the initial grasping points (black dots) and the step order (numbers) on the graspability-assessed terrain map. The 7th step represents a regrasping action around the same foothold. Regions with very low graspability are omitted from the visualization.}
  \label{fig:limbero_climbing_exp}
\end{figure*}
\begin{figure}[t]
  \centering
  \includegraphics[width=\linewidth]{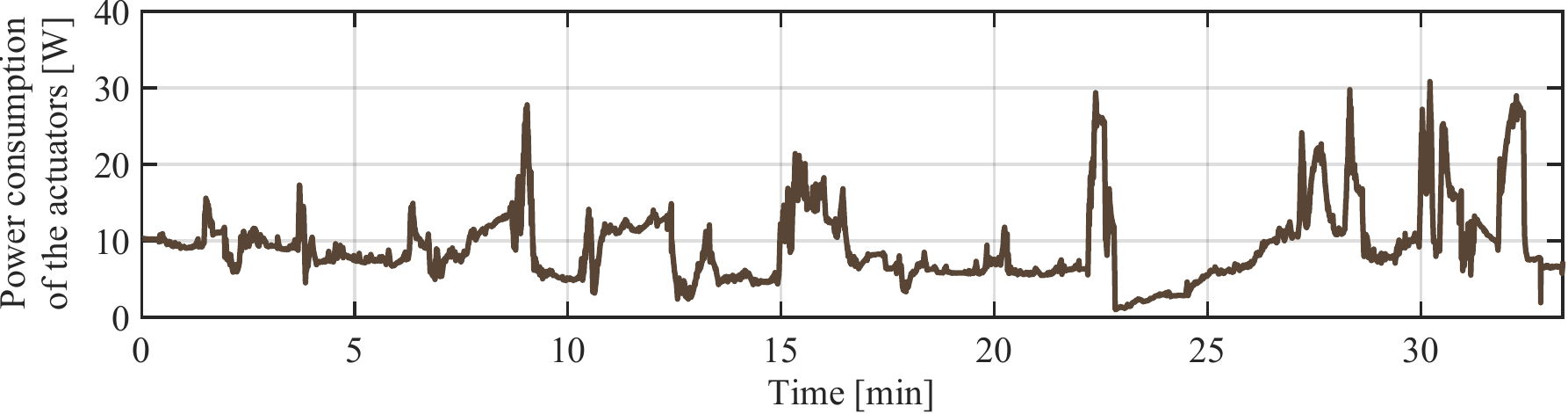} 
  \caption{Power consumption of all actuators of LIMBERO during climbing. Several peaks correspond to periods when the robot pushes the base upward.}
  \label{fig:power_exp}
\end{figure}
Climbing experiments were conducted to demonstrate LIMBERO's rock climbing capability. In these experiments, the robot was manually teleoperated by a human operator. The operator planned the foothold sequence based on the pre-obtained graspability-assessed terrain map. The teleoperation commands included selecting the next swing limb, specifying a reference velocity for the limb, and issuing gripping and releasing commands.


LIMBERO successfully climbed an analogous rocky terrain under full Earth gravity, which has not yet been achieved by 10\;kg-class limbed robots of this type. The climbing experiment was repeated ten times, and \fig{fig:limbero_climbing_exp} shows one representative trial. Across the ten trials, the robot completed a total of 44 steps, with a mean time per step of 3.8\;min. 
On average, the robot ascended 0.5\;m in 30\;min, corresponding to a climbing speed of 0.017\;m/s. One major reason for this relatively slow velocity is the additional time required for manual adjustment of the gripper position, which could be reduced through future autonomous limb control. The total power consumption of the actuators during climbing averaged 10\;W, with a range of 5--30\;W (\fig{fig:power_exp}). We confirmed that higher joint torques were required to move the robot base during the supporting phase, as evidenced by several peaks in the time history.

We occasionally encountered instances in which the shoulder and wrist joints exceeded their torque limits, requiring a reboot command to restore their functionality. In some cases, the testing was halted for safety. We also observed that the supporting limbs' grippers intermittently slipped while scratching the rock surface during upward base motion. In most cases, the grippers still maintained their grasping state; only one unintended detachment occurred across the ten trials. Nevertheless, this preliminary experiment demonstrated LIMBERO's potential for traversing rocky cliff terrain, and all the critical lessons learned obtained throughout these tests are summarized in the next section.

\subsection{Lessons Learned}
While LIMBERO demonstrated stable holding performance and climbing capability on rocky cliffs under full Earth gravity, further hardware improvements are necessary for more stable locomotion. It is noteworthy that the developed spine gripper is more than sufficient to stably support the robot and a payload exceeding 10\;kg. However, significant torque is required to move the robot base upward, highlighting the need for enhanced shoulder joints. Moreover, for limbed climbing robots in particular, the closed mechanical chain formed by limbs strongly adhered to the environment via grippers can result in concentrated torque on certain joints, which can also lead to exceeding torque limits. To mitigate this unbalanced internal force distribution, the incorporation of mechanical flexibility, reaction mitigating motion planning~\cite{warley2023ramp} or force-accommodating control strategies (e.g., impedance control)~\cite{imai2024admittance} would play a vital role.

It is also noteworthy that the spine tips were slightly worn after repeated use (more than 100 grasping cycles), even though high-durability, vacuum-hardened spikes were employed. However, no noticeable decline in grasping performance was observed on the lava rocks used in this study.

Furthermore, although the proposed algorithm can reasonably evaluate the graspability score across the terrain geometry, additional autonomy is essential to prevent unexpected or unmeasurable factors from causing gripper detachment (e.g., terrain surface breakage). Therefore, a practical scenario for the rock-climbing robot can be outlined as follows: the robot first scans and maps the terrain in three dimensions, detects the graspable geometries in the map, and then haptically assesses each part to ensure it is neither slippery, broken, nor excessively soft. 


\section{Conclusion}
In this paper, we developed LIMBERO, a quadrupedal climbing robot designed to achieve enhanced cliff-gripping and climbing capabilities for the practical deployment of limbed climbing robots. LIMBERO’s novel active spine gripper, featuring coupled biting and hooking motions, demonstrated exceptional performance in stable rocky cliff-hanging and climbing under 1\;G conditions in real-world experiments. Furthermore, we proposed an efficient algorithm to extract graspable geometries from 3D terrain maps, enabling effective terrain assessment for climbing tasks. 

Lessons learned from these experiments underscore the need for further enhancement of the limb joints to improve upward movement on vertical surfaces, which is crucial for the practical deployment of LIMBERO not only in planetary exploration but also in terrestrial applications. These outcomes lay a significant foundation for the future development of more capable and autonomous limbed climbing robots. 

In addition to further hardware sophistication, future work includes more autonomous operation exploiting optimal foothold selection among the discrete graspable points proposed in~\cite{Takada_ICRAE2023}, and efficient control methods for reducing contact reaction forces and balancing joint torque distribution during climbing, as studied in~\cite{imai2024admittance,warley2023ramp}, while maintaining the tumble stability criteria proposed in~\cite{yonedaIROS1996tumble,ribeiro2020dynamic}. 

\section*{Acknowledgment}
The authors would like to thank Mao Kasano, Kohta Naoki, Koki Murase, Takuya Kato, Jitao Zheng, Keigo Haji, and Taku Okawara for their contributions to the productive discussions and their assistance in the development of the robot system. The authors also express their sincere appreciation to the Mechanical Workshop Section 2, Technical Division, Tohoku University, for their support in fabricating the mechanical components.

\bibliography{./IEEEabrv,bibliography.bib}

\end{document}